\documentclass[11pt,a4paper]{article}
\usepackage[hyperref]{naaclhlt2019}
\usepackage{times}
\usepackage{latexsym}

\usepackage{hyperref}
\usepackage{url}
\usepackage{booktabs}

\usepackage{outlines}
\usepackage{amsmath,amsthm,amssymb}
\usepackage{bm}
\usepackage{graphicx}
\usepackage{caption}
\usepackage{subcaption}
\usepackage{natbib}
\usepackage{mathtools}
\usepackage{soul}
\usepackage{xcolor}
\usepackage{multirow}
\usepackage{url}
\usepackage{bbm}
\usepackage{cleveref}
\usepackage{tabularx}
\usepackage{tabulary}

\usepackage{footnote}

\usepackage{pifont}
\newcommand{\cmark}{\ding{51}}%

\aclfinalcopy 
\def\aclpaperid{829} 

\title{Learning to Attend On Essential Terms: An Enhanced Retriever-Reader Model for Open-domain  Question Answering}

\author{Jianmo Ni$^{1}$\thanks{~Most of the work was done during internship at Microsoft, Redmond.}, Chenguang Zhu$^2$,  Weizhu Chen$^3$,  Julian McAuley$^1$  \\
$^1$ University of California, San Diego \\
$^2$ Microsoft Speech and Dialogue Research Group \\ 
$^3$ Microsoft Dynamics 365 AI \\
\texttt{\{jin018,jmcauley\}@ucsd.edu, \{chezhu,wzchen\}@microsoft.com} \\
}

\date{}

\begin{document}
\maketitle
\begin{abstract}
Open-domain question answering remains a challenging task as it requires models that are capable of understanding questions and answers, collecting useful information, and reasoning over evidence. Previous work typically formulates this task as a reading comprehension or entailment problem given evidence retrieved from search engines. However, existing techniques struggle to retrieve indirectly related evidence when no directly related evidence is provided, especially for complex questions where it is hard to parse precisely what the question asks. In this paper we propose a retriever-reader model that learns to attend on essential terms during the question answering process. We build (1) an essential term selector which first identifies the most important words in a question, then reformulates the query and searches for related evidence; and (2) an enhanced reader that distinguishes between essential terms and distracting words to predict the answer. We evaluate our model on multiple open-domain multiple-choice QA datasets, notably performing at the level of the state-of-the-art on the AI2 Reasoning Challenge (ARC) dataset.
\end{abstract}

\section{Introduction}

Open-domain question answering (QA) has been extensively studied in recent years. Many existing works have followed the `search-and-answer' strategy and achieved strong performance \citep{Chen2017ReadingWT, Kwon2018ControllingIA,Wang2018R3RR}
spanning multiple QA datasets such as TriviaQA \citep{Joshi2017TriviaQAAL}, SQuAD \citep{Rajpurkar2016SQuAD10}, MS-Macro \citep{Nguyen2016MSMA}, ARC \citep{Clark2018ThinkYH} among others. 

However, open-domain QA tasks become inherently more difficult
when (1) dealing with questions with little available evidence;
(2) solving questions where the answer type is free-form text (e.g.~multiple-choice) rather than a span among existing passages (i.e.,~`answer span'); or when (3) the need arises to understand long and complex questions and reason over multiple passages, rather than simple text matching.
As a result, it is essential to incorporate commonsense knowledge or to improve retrieval capability to better capture
partially related evidence
\citep{Chen2017ReadingWT}. 

As shown in \Cref{tbl:diff}, the TriviaQA, SQuAD, and MS-Macro datasets all provide passages within which the correct answer is guaranteed to exist. 
However, this assumption ignores the difficulty of retrieving question-related evidence from a large volume of open-domain resources, especially when considering complex questions which require reasoning or commonsense knowledge. 
On the other hand, ARC does not provide passages known to contain the correct answer. Instead, the task of identifying relevant passages is left to the solver. 
However, questions in ARC have multiple answer choices that provide indirect information that can help solve the question. 
As such an effective model needs to account for relations among passages, questions, and answer choices. Real-world datasets such as Amazon-QA (a corpus of user queries from Amazon) \citep{McAuley2016AddressingCA} also exhibit the same challenge, i.e., the need to surface related evidence from which to extract or summarize an answer.

\begin{savenotes}
\begin{table*}[t]\fontsize{9}{11}\selectfont
\centering
\vspace{-0.5\baselineskip}
\begin{tabular}{lrrrrr}
\toprule
Dataset & \# of questions & \parbox{0.08\textwidth}{\centering Open-domain} & \parbox{0.08\textwidth}{\centering Multiple choice} & \parbox{0.08\textwidth}{\centering Passage retrieval} & \parbox{0.12\textwidth}{\centering No ranking supervision\footnote{For SQuAD and TriviaQA, since the questions are paired with span-type answers, it is convenient to obtain ranking supervision where retrieved passages are relevant via distant supervision; however free-form questions in ARC and Amazon-QA result in a lack of supervision which makes the problem more difficult. For MS-Macro, the dataset is designed to annotate relevant passages though it has free-form answers.}} \\
\midrule
ARC \citep{Clark2018ThinkYH} &  $\approx$ 7K & \cmark & \cmark  & \cmark & \cmark \\ 
Amazon-QA \citep{McAuley2016AddressingCA}  & $\approx$ 1.48M  & \cmark &   &    &  \cmark \\ \midrule
SQuAD \citep{Rajpurkar2016SQuAD10} & $\approx$ 100K &  \cmark &    &   &   \\ 
TriviaQA \citep{Joshi2017TriviaQAAL} & $\approx$ 650K &  \cmark &    &   &   \\ 
MS-Macro \citep{Nguyen2016MSMA} &  $\approx$ 1M &  \cmark &   &   &   \\
\bottomrule
\end{tabular}
\caption{Differences among popular QA datasets. 
\label{tbl:diff}}
\end{table*}
\end{savenotes}

\begin{figure*}[t]
    \centering
     \includegraphics[height=2.6in]{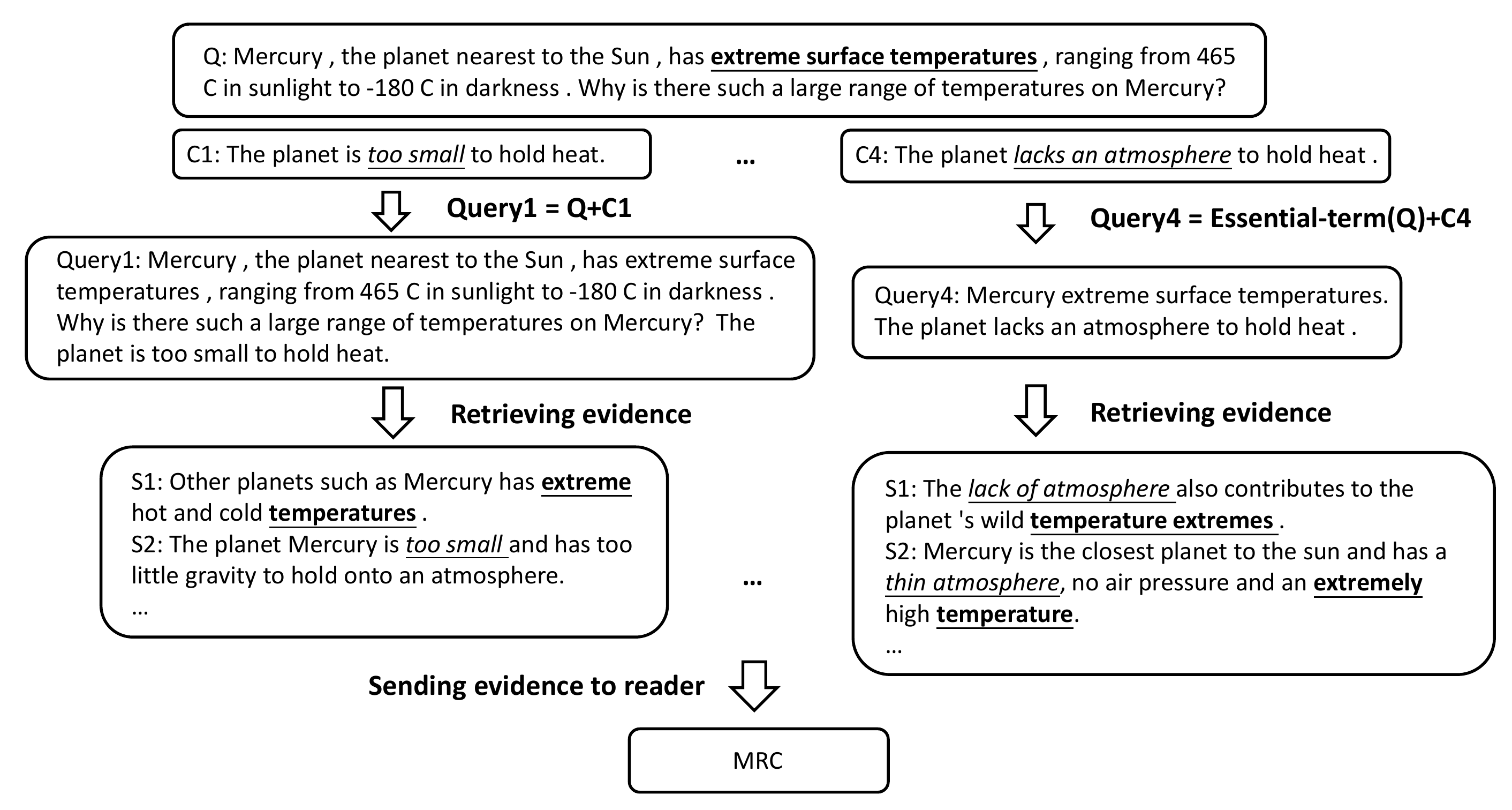}
    \caption{Example of the retrieve-and-read process to solve open-domain questions. Words related with the question are in \underline{\textbf{bold}}; and words related with C1 and C4 are in \underline{\textit{italics}}. 
    \label{fig:motivation}
    }
\end{figure*}

\Cref{fig:motivation} shows an example of a question in the ARC dataset and demonstrates the difficulties in retrieval and reading comprehension. As shown for Choice 1 (C1), a simple concatenation of the question and the answer choice is not a reliable query and is of little help when trying to find supporting evidence to answer the question (e.g.~we might retrieve sentences similar to the question or the answer choice, but would struggle to find evidence explaining
\emph{why} the answer choice is correct). On the other hand, a reformulated query consisting of essential terms in the question and Choice 4 can help retrieve evidence explaining why Choice 4 is a correct answer. To achieve this, the model needs to 
(1) ensure that the retrieved evidence supports the fact mentioned in both the question and the answer choices and (2) capture this information and predict the correct answer.

To address these difficulties, we propose an essential-term-aware Retriever-Reader (ET-RR) model that learns to attend on essential terms during retrieval and reading. Specifically, we develop a two-stage method with an essential term selector followed by an attention-enhanced reader.

\textbf{Essential term selector.} ET-Net is a recurrent neural network that seeks
to understand the question and select essential terms, i.e.,~key words, from the question. We frame this problem as a classification task for each word in the question. These essential terms are then concatenated with each 
answer choice 
and fed into a retrieval engine to obtain related evidence.

\textbf{Attention-Enhanced Reader.} Our neural reader takes the triples (question, answer choice, retrieved passage) as input. 
The reader consists of a sequence of language understanding layers: an input layer, attention layer, sequence modeling layer, fusion layer, and an output layer. 
The attention and fusion layers help the model to obtain a refined representation of one text sequence based on the understanding of another, e.g.~a passage representation based on an understanding of the question. We further add a choice-interaction module to handle the semantic relations and differences between answer choices. Experiments show that this can further improve the model's accuracy. 

We evaluate our model on the ARC Challenge dataset, where our model achieves an accuracy of 36.61\% on the test set, 
and outperformed all leaderboard solutions 
at the time of 
writing (Sep.~2018).
To compare with other benchmark datasets, we adapt RACE \citep{lai2017large} and MCScript \citep{Ostermann2018SemEval2018T1} to the open domain setting by removing their
supervision in the form of relevant passages. We also consider a large-scale real-world open-domain dataset, Amazon-QA, to evaluate our model's scalability and to compare against standard benchmarks designed for the open-domain setting. Experiments on these three datasets show that ET-RR outperforms baseline models by a large margin.
We conduct multiple ablation studies to show the effectiveness of each component of our model.
Finally, we perform in-depth error analysis to explore the model's limitations.

\section{Related Work}

There has recently been growing interest in building better retrievers for open-domain QA. 
\citet{Wang2018R3RR} proposed a Reinforced Ranker-Reader model that ranks  retrieved evidence and assigns different weights to evidence prior to processing by the reader.
\citet{Min2018EfficientAR} demonstrated that for several popular MRC datasets (e.g.~SQuAD, TriviaQA) most questions can be answered using only a few sentences rather than the entire document. Motivated by this observation, they built a sentence selector to gather this potential evidence for use by the reader model.
\citet{Nishida2018RetrieveandReadML} developed a multi-task learning (MTL) method for a retriever and reader in order to obtain a strong retriever that considers certain passages including the answer text as positive samples during training. 
The proposed MTL framework is still limited to scenarios where it is feasible to discover whether the passages contain the answer span.
Although these works have achieved progress
on open-domain QA by improving the ranking or selection of given evidence, few have focused on 
the scenario where the model needs to start by searching for the evidence itself. 

Scientific Question Answering (SQA) is a representative open-domain task that requires capability in both retrieval and reading comprehension. In this paper, we study question answering on the AI2 Reasoning Challenge (ARC) scientific QA dataset \citep{Clark2018ThinkYH}.
This dataset contains multiple-choice scientific questions from 3rd to 9th grade standardized tests and a large corpus of relevant information gathered from search engines.
The dataset is partitioned into  ``Challenge'' and ``Easy'' sets. The challenge set consists of questions that cannot be answered correctly by either of the solvers based on Pointwise Mutual Information (PMI) or Information Retrieval (IR). 
Existing models tend to achieve only slightly better and sometimes even worse performance than random guessing, which shows that existing models are not well suited to this kind of QA task. 

\citet{Jansen2017FramingQA} first developed a rule-based focus word extractor to identify essential terms in the question and answer candidates. The extracted terms are used to aggregate a list of potential answer justifications for each answer candidate. Experiments shown that focus words are beneficial for SQA on a subset of the ARC dataset. \citet{Khashabi2017LearningWI} also worked on the problem of finding essential terms in a question for solving SQA problems. They published a dataset containing over 2,200 science questions annotated with essential terms and train multiple classifiers on it. Similarly, we leverage this dataset to build an essential term selector using a neural network-based algorithm.
More recently, \citet{Boratko2018ASC} developed a labeling interface to obtain high quality labels for the ARC dataset. One finding is that human annotators tend to retrieve better evidence after they reformulate the search queries which are originally constructed by a simple concatenation of question and answer choice.
By feeding the evidence obtained by human-reformulated queries into a pre-trained MRC model (i.e.,~DrQA \citep{Chen2017ReadingWT}) 
they achieved an accuracy increase of 42\% on a subset of 47 questions. This shows the potential for a ``human-like'' retriever to boost performance on this task. 

Query reformulation has been shown to be effective in information retrieval \citep{Lavrenko2001RelevanceBasedLM}. \citet{Nogueira2017TaskOrientedQR} modeled the query reformulation task as a binary term selection problem (i.e.,~whether to choose the term in the original query and the documents retrieved using the original query). The selected terms are then concatenated to form the new query. Instead of choosing relevant words, \citet{Buck2018AskTR} proposed a sequence-to-sequence model to generate new queries.  
\citet{das2018multistep} proposed Multi-step Retriever-Reader  which explores an iterative retrieve-and-read strategy for open-domain question answering. It formulates the query reformulation problem in the embedding space where the vector representation of the question is changed to improve the performance. Since there is no supervision for training the query reformulator, all these methods using reinforcement learning to maximize the task-specific metrics (e.g.~Recall for paragraph ranking, F1 and Exact Matching for span-based MRC). Different from these works, we train the query reformulator using an annotated dataset as supervision and then apply the output to a separate reader model. We leave the exploration of training our model end-to-end using reinforcement learning as future work.

\begin{figure*}[t]
    \centering
     \includegraphics[height=2.5in]{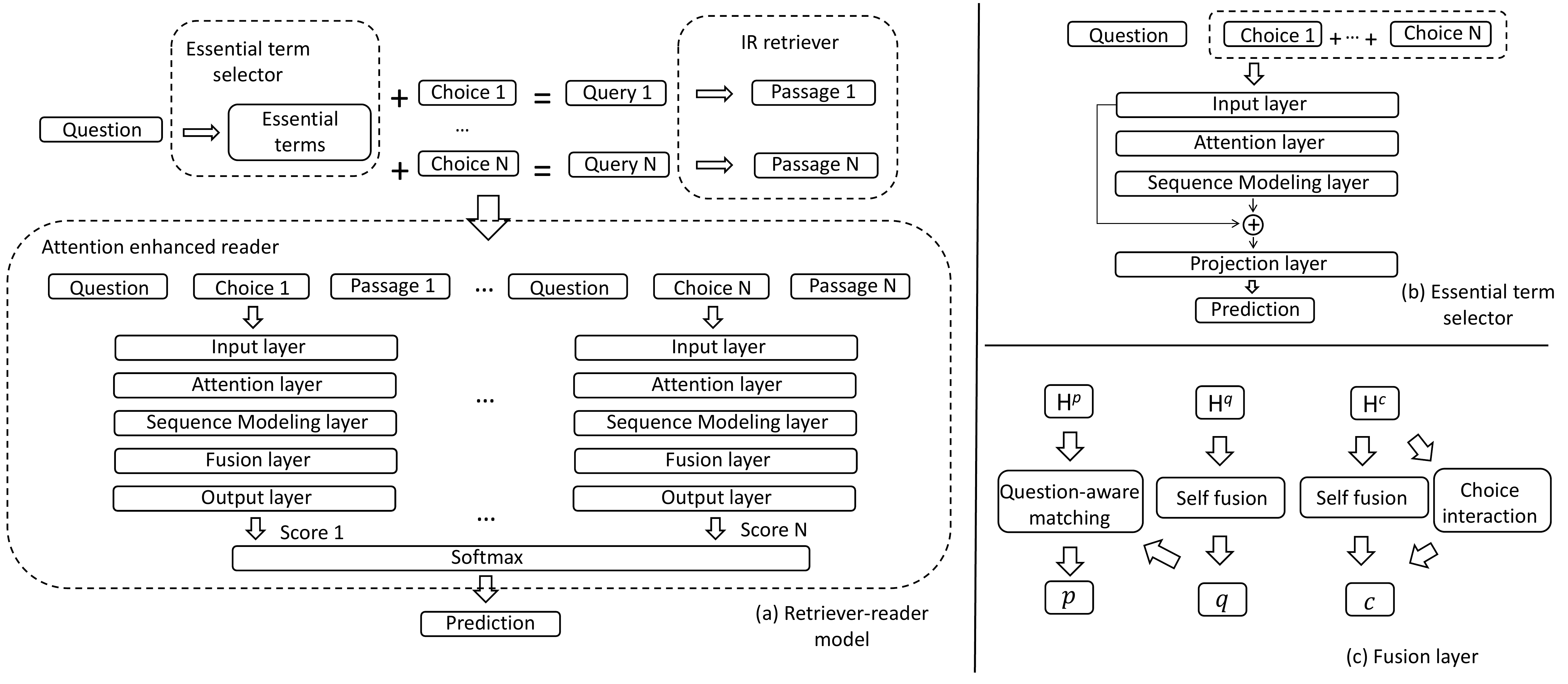}
    \caption{Model structure for our essential-term-aware retriever-reader model.
    \vspace{-0.5\baselineskip}
    \label{fig:model}
    }
\end{figure*}

\section{Approach}
In this section, we introduce the essential-term-aware retriever-reader model (ET-RR). 
As shown in \Cref{fig:model}, we build a term selector to discover which terms are essential in a question. The selected terms are then used to formulate a more efficient query enabling the retriever to obtain related evidence. The retrieved evidence is
then fed to the reader to predict the final answer.

For a question with $q$ words $\mathbf{Q} = \{w_t^Q\}_{t=1}^q$ along with its $N$ answer choices $\mathbf{C} = \{\mathbf{C}_n\}_{n=1}^{N}$ where $\mathbf{C}_n=\{w_t^C\}_{t=1}^c$,  
the essential-term selector chooses a subset of essential terms $\mathbf{E}\subset\mathbf{Q}$, 
which are then concatenated with
each $\mathbf{C}_n$ to formulate a query. The query for each answer choice, $\mathbf{E}+\mathbf{C}_n$, is sent to the retriever (e.g.~Elastic Search\footnote{https://www.elastic.co/products/elasticsearch}), and the top $K$ retrieved sentences based on the scores returned by the retriever are then concatenated into the evidence passage $\mathbf{P}_n=\{w_t^P\}_{t=1}^K$.

Next, given these text sequences $\mathbf{Q}$, $\mathbf{C}$, and $\mathbf{P} = \{\mathbf{P}_n\}_{n=1}^{N}$, the reader will determine
a matching score for each triple $\{\mathbf{Q}, \mathbf{C}_n, \mathbf{P}_n\}$.
The answer choice $\mathbf{C}_{n^*}$ with the highest score is selected.

We first introduce the reader model in \Cref{sec:reader} and then the essential term selector in \Cref{sec:et}.

\subsection{Reader Model}
\label{sec:reader}
\subsubsection{Input Layer}
\label{sec:inputlayer}
To simplify notation, we ignore the subscript $n$ denoting the answer choice until the final output layer. In the input layer, all text inputs---the question, answer choices, and passages, i.e.,~retrieved evidence---are converted into 
embedded representations. Similar to \citet{Wang2018YuanfudaoAS}, we consider the following components for each word:  

\noindent\textbf{Word Embedding}. Pre-trained GloVe word embedding with dimensionality $d_w=300$.

\noindent\textbf{Part-of-Speech Embedding and 
Named-Entity Embedding}. The part-of-speech tags and named entities for each word are mapped to embeddings with dimension 16.

\noindent\textbf{Relation Embedding}. A relation between each word in $\mathbf{P}$ and any word in $\mathbf{Q}$ or $\mathbf{C}$ is mapped to an embedding with dimension 10. In the case that multiple relations exist, we select one uniformly at random. The relation is obtained by querying ConceptNet \citep{Speer2017ConceptNet5A}.

\noindent\textbf{Feature Embeddings}. 
Three handcrafted features are used to enhance the word representations:
(1) Word Match; if a word or its lemma of $\mathbf{P}$ exists in $\mathbf{Q}$ or $\mathbf{C}$, then this feature is 1 (0 otherwise).
(2) Word Frequency; a logarithmic term frequency is calculated for each word.
(3) Essential Term; for the $i$-th word in $\mathbf{Q}$, this feature, denoted as $w_{e_i}$, is 1 if the word is an essential term (0 otherwise). Let $\mathbf{w}_e=[w_{e_1}, w_{e_2}, \ldots, w_{e_q}]$ denote the essential term vector.

For $\mathbf{Q}, \mathbf{C}, \mathbf{P}$, all of these components are concatenated to obtain the final word representations $\mathbf{W}_Q \in \mathbb{R}^{q \times d_{Q}}, \mathbf{W}_C \in \mathbb{R}^{c \times d_{C}}, \mathbf{W}_P \in \mathbb{R}^{p \times d_{P}}$, where $d_{Q}, d_{C}, d_{P}$ are the final word dimensions of $\mathbf{Q}, \mathbf{C},$ and $\mathbf{P}$.

\subsubsection{Attention Layer}
As shown in \Cref{fig:model}, after obtaining word-level embeddings, attention is added to enhance word representations. 
Given two word embedding sequences $\mathbf{W}_U, \mathbf{W}_V$, 
word-level attention is calculated as:
\begin{equation}
  \begin{aligned} \label{eq:seqattn}
  \mathbf{M}^{'}_{UV} & = \mathbf{W}_U\mathbf{U} \cdot (\mathbf{W}_V\mathbf{V})^{\top}  \\
  \mathbf{M}_{UV} & = \mathrm{softmax}(\mathbf{M}^{'}_{UV}) \\
  \mathbf{W}_U^V & = \mathbf{M}_{UV} \cdot (\mathbf{W}_V\mathbf{V}),
  \end{aligned}
\end{equation}
where $\mathbf{U} \in \mathbb{R}^{d_{U} \times d_{w}}$ and $\mathbf{V} \in \mathbb{R}^{d_{V} \times d_{w}}$ are two matrices that convert word embedding sequences to dimension $d_{w}$, and $\mathbf{M}^{'}_{UV}$
contains dot products between each word in $\mathbf{W}_U$ and $\mathbf{W}_V$, and softmax is applied on $\mathbf{M}^{'}_{UV}$ 
row-wise. 
Three types of attention are calculated using \Cref{eq:seqattn}:
(1) question-aware passage representation $\mathbf{W}_P^Q \in \mathbb{R}^{p \times d_w}$;
(2) question-aware choice representation $\mathbf{W}_C^Q \in \mathbb{R}^{c \times d_w}$; and 
(3) passage-aware choice representation $\mathbf{W}_C^P \in \mathbb{R}^{c \times d_w}$.

\subsubsection{Sequence Modeling Layer}
To model the contextual dependency of each text sequence, we 
use BiLSTMs to process the word representations obtained from the input layer and attention layer:
\begin{equation}
  \begin{aligned} \label{eq:rnn}
  \mathbf{H}^{q} & = \mathbf{BiLSTM}{[\mathbf{W}_Q]} \\
  \mathbf{H}^{c} & = \mathbf{BiLSTM}{[\mathbf{W}_C; \mathbf{W}_C^P; \mathbf{W}_C^Q]} \\
  \mathbf{H}^{p} & = \mathbf{BiLSTM}{[\mathbf{W}_P; \mathbf{W}_P^Q]},
  \end{aligned}
\end{equation}
where $\mathbf{H}^{q} \in \mathbb{R}^{q \times l}, \mathbf{H}^{c} \in \mathbb{R}^{c \times l}$, and $\mathbf{H}^{p} \in \mathbb{R}^{p \times l}$ are the hidden states of the BiLSTMs, `;' is feature-wise concatenation, and $l$ is the size of the hidden states.

\subsubsection{Fusion Layer}
We further convert each question and answer choice into a single vector: $\mathbf{q} \in \mathbb{R}^{l}$ and $\mathbf{c} \in \mathbb{R}^{l}$:
\begin{equation}
  \begin{aligned} \label{eq:selfattn}
  \bm{\alpha}_q & = \mathrm{softmax}([\mathbf{H}^{q}; \mathbf{w}_{e}] \cdot \mathbf{w}_{sq}^{\top}),~\mathbf{q} = {\mathbf{H}^{q}}^{\top}\bm{\alpha}_q \\
  \bm{\alpha}_c & = \mathrm{softmax}(\mathbf{H}^{c} \cdot \mathbf{w}_{sc}^{\top}),~\mathbf{c} = {\mathbf{H}^{c}}^{\top}\bm{\alpha}_c,
  \end{aligned}
\end{equation}
where the essential-term feature $\mathbf{w}_{e}$ from \Cref{sec:inputlayer} is concatenated with $\mathbf{H}^{q}$, and $\mathbf{w}_{sq}$ and $\mathbf{w}_{sc}$ are learned parameters.

Finally, a bilinear sequence matching is calculated between $\mathbf{H}^{p}$ and $\mathbf{q}$ to obtain a question-aware passage representation, which is used as the final passage representation:
\begin{equation}
  \label{eq:pqattn}
  \bm{\alpha}_p = \mathrm{softmax}(\mathbf{H}^{p} \cdot \mathbf{q}); \quad
  \mathbf{p}  = {\mathbf{H}^{p}}^{\top}\bm{\alpha}_p.
\end{equation}

\subsubsection{Choice Interaction}
When a QA task provides multiple choices for selection, the relationship between the choices can provide useful information to answer the question. Therefore, we integrate a choice interaction layer to handle the semantic correlation between multiple answer choices. 
Given the hidden state $\mathbf{H}^{c_n}$ of choice $c_n$ and $\mathbf{H}^{c_i}$ of other choices $c_i$, $\forall i \neq n$, we calculate the differences between the hidden states and apply max-pooling over the differences:
\begin{equation}
  \mathbf{c}_{\mathit{inter}} = \mathrm{Maxpool} (\mathbf{H}^{c_n} - \frac{1}{N-1} \sum_{i \neq n} \mathbf{H}^{c_i}),
\end{equation}
where $N$ is the total number of answer choices. Here, $\mathbf{c}_{\mathit{inter}}$ characterizes the differences between an answer choice $c_n$ and other answer choices. The final representation of an answer choice is updated by concatenating the self-attentive answer choice vector and inter-choice representation as $\mathbf{c}^{\mathit{final}} = [\mathbf{c}; \mathbf{c}_{\mathit{inter}}].$

\subsubsection{Output Layer}
For each tuple $\{\mathbf{q}, \mathbf{p}_n, \mathbf{c}_n\}_{n=1}^{N}$, two scores are calculated by matching 
(1)
the passage and answer choice and
(2) question and answer choice. We use a bilinear form for both matchings. Finally, a softmax function is applied over $N$ answer choices to determine the best answer choice: 
\begin{equation}
  \begin{aligned} 
  \label{eq:pred}
  s_n^{pc} = \mathbf{p}_n \mathbf{W}^{pc} \mathbf{c}^{\mathit{final}}_n 
  ; \quad 
  s_n^{qc} = \mathbf{q} \mathbf{W}^{qc} \mathbf{c}^{\mathit{final}}_n \\
  \mathbf{s} = \mathrm{softmax}(\mathbf{s}^{pc}) + \mathrm{softmax}(\mathbf{s}^{qc}),
  \end{aligned}
\end{equation}
where $s_n^{pc}, s_n^{qc}$ are the scores for answer choice $1 \leq n \leq N$; $\mathbf{s}^{pc}, \mathbf{s}^{qc}$ are score vectors for all $N$ choices; and $\mathbf{s}$ 
contains the final scores for each answer choice. During training, we use a cross-entropy loss.

\begin{table*}
\parbox{.45\linewidth}{
\centering
\vspace{-0.5\baselineskip}
\setlength{\tabcolsep}{2pt}
\begin{tabularx}{\linewidth}{lX} 
\toprule
\small{Question}    & \small{If an object is attracted to a magnet, the object is most likely made of (A) wood (B) plastic (C) cardboard (D) metal} \\ \hline
\small{\# annotators} & \small{5} \\ \hline
\small{Annotation}  &  \small{If,0; an,0; object,3; is,0; attracted,5; to,0; a,0; 
magnet,,5; the,0; object,1; is,0; most,0; likely,0; made,2; of,0} \\ 
\bottomrule
\end{tabularx}
\setlength{\tabcolsep}{6pt}
\caption{Example of essential term data.\label{tbl:selector-dataset}}
}
\hfill
\parbox{.45\linewidth}{
\fontsize{9}{11}\selectfont
\centering
\vspace{-0.5\baselineskip}
\begin{tabular}{lccc}
\toprule
Model &  Precision & Recall & F1 \\    
\midrule
MaxPMI        & 0.88      & 0.65   & 0.75  \\
SumPMI        & 0.88      & 0.65   & 0.75  \\
PropSurf      & 0.68      & 0.64   & 0.66  \\
PropLem       & 0.76      & 0.64   & 0.69  \\
ET Classifier & 0.91      & 0.71   & 0.80 \\
ET-Net         & 0.74     & \textbf{0.90}  & \textbf{0.81} \\
\bottomrule
\end{tabular}
\caption{
Performance of different selectors.\label{tbl:selector-result}}
}
\end{table*}

\subsection{Essential Term Selector}
\label{sec:et}
Essential terms are key words in a question that are crucial in helping a retriever obtain related evidence. Given a question $\mathbf{Q}$ and $N$ answer choices $\mathbf{C}_1,\ldots,\mathbf{C}_N$,
the goal is to predict a binary variable $y_i$ for each word $Q_i$ in the question $\mathbf{Q}$, where $y_i=1$ if $Q_i$ is an essential term and 0 otherwise. 
To address this problem, we build a neural model, ET-Net, which has the same design as the reader model for the input layer, attention layer, and sequence modeling layer to obtain the hidden state $\mathbf{H}^q$ for question $\mathbf{Q}$.

In detail, we take the question $\mathbf{Q}$ and the concatenation  $\mathbf{C}$ of all $N$ answer choices as input to ET-Net.
$\mathbf{Q}$ and $\mathbf{C}$ first go through an input layer to convert to the embedded word representation, and then word-level attention is calculated to obtain a choice-aware question representation $\mathbf{W}_Q^C$ as in \Cref{eq:seqattn}.
We concatenate the word representation and word-level attention representation of the question and feed it into the sequence modeling layer: 
\begin{equation}
  \label{eq:et-rnn}
  \mathbf{H}^{q} = \mathbf{BiLSTM}{[\mathbf{W}_Q; \mathbf{W}_Q^C]}.
\end{equation}

As shown in \Cref{fig:model},
the hidden states obtained from the attention layer are then concatenated with the embedded representations of $\mathbf{Q}$ and fed into a projection layer to obtain the prediction vector $\mathbf{y} \in \mathbb{R}^{q}$ for all words in the question:
\begin{equation}
\mathbf{y} = [\mathbf{H}^q; \mathbf{W}_Q^{f}] \cdot \mathbf{w}^s,
\end{equation}
where $\mathbf{w}^s$ contains the learned parameters, and $\mathbf{W}_Q^{f}$ is the concatenation of the POS embedding, NER embedding, relation embedding, and feature embedding from \Cref{sec:inputlayer}.

After obtaining the prediction for each word, we use a binary cross-entropy loss to train the model. During evaluation, we take words with $y_i$ greater than 0.5 as essential terms.

\section{Experiments}

In this section, we first discuss the performance of the essential term selector, ET-Net, on a public dataset. We then discuss the performance of the whole retriever-reader pipeline, ET-RR, on multiple open-domain datasets.
For both the ET-Net and ET-RR models, we use 96-dimensional hidden states and 1-layer BiLSTMs in the sequence modeling layer. A dropout rate of 0.4 is applied for the embedding layer and the BiLSTMs' output layer. We use adamax \citep{Kingma2014AdamAM} with a learning rate of 0.02 and batch size of 32. The model is trained for 100 epochs. 
Our code is released at \url{https://github.com/nijianmo/arc-etrr-code}.

\subsection{Performance on Essential Term Selection}
We use the public dataset from \citet{Khashabi2017LearningWI} which contains 2,223 annotated questions, each accompanied by four answer choices. \Cref{tbl:selector-dataset} gives an example of an annotated question. As shown, the dataset is annotated for binary classification. For each word in the question, the data measures
whether the word is an ``essential'' term according to 5 annotators. 
We then split the dataset into training, development, and test sets using an  8:1:1 ratio and select the model that performs best on the development set. 

\Cref{tbl:selector-result} shows the performance of our essential term selector and baseline models from \citet{Khashabi2017LearningWI}.
The second best model (ET Classifier) is an SVM-based model from \citet{Khashabi2017LearningWI} requiring over 100 handcrafted features. 
As shown, our ET-Net achieves a comparable result with ET Classifier in terms of the F1 Score.

\Cref{tbl:selector-example} shows example predictions made by ET-Net. As shown, ET-Net is capable of selecting most ground-truth essential terms. It rejects certain words such as ``organisms'' which have a high TF-IDF in the corpus but are not relevant to answering a particular question. This shows its ability to discover essential terms according to the context of the question.

\begin{table*}
\parbox{.5\linewidth}{
\fontsize{9}{11}\selectfont
\centering
\vspace{-0.5\baselineskip}
\begin{tabularx}{\linewidth}{X}
\toprule
Example questions \\
\midrule
Which \textbf{\underline{unit}} of \textbf{\underline{measurement}} can be used to describe the \textbf{\underline{length}} of a \textbf{\underline{desk}} ? \\
One way \textbf{\underline{animal}} usually \textbf{\underline{respond}} to a sudden \textbf{\underline{drop}} in \textbf{\underline{temperature}} is by \\
Organisms require \underline{energy} to \underline{survive}. Which of the following \textbf{\underline{processes provides energy}} to the \underline{body} ? \\
\bottomrule
\end{tabularx}
\caption{Examples of essential term prediction (in questions) by ET-Net. True positives are marked bold and underlined while false positives are only underlined.\label{tbl:selector-example}}
}
\hfill
\parbox{.45\linewidth}{
\fontsize{9}{11}\selectfont
\centering
\vspace{-0.5\baselineskip}
\begin{tabular}{lrrrr}
\toprule
Dataset & Train & Dev & Test & Corpus \\ \midrule
ARC & 1,119 & 299 & 1,172 & 1.46M \\ 
RACE-Open & 9,531 & 473 & 528 & 0.52M \\ 
MCScript-Open &  1,036 & 156 & 319 & 24.2K \\ 
Amazon-Patio & 36,587  &  4,531  &    4,515 &  2.55M   \\
Amazon-Auto &  49,643  &  6,205  &  6,206   &  7.32M   \\
Amazon-Cell & 40,842  &  5,105   &     5,106  & 1.86M    \\
\bottomrule
\end{tabular}
\caption{Statistics on ARC, RACE-Open, MCScript-Open and Amazon-QA. Corpus size is the number of sentences. \label{tbl:datasets}}
}
\end{table*}

\begin{table*}
\fontsize{9}{11}\selectfont
\centering
\vspace{-0.5\baselineskip}
\begin{tabularx}{\linewidth}{lX} 
\toprule
Dataset                    & Example questions     \\ \midrule
\multirow{2}{*}{ARC}       & The best way to \underline{separate salt} from \underline{water} is with the use of     \\
                           & Which \underline{geologic} process most likely \underline{caused} the \underline{formation} of the Mount St. \underline{Helens Volcano}? \\ \midrule
\multirow{2}{*}{RACE-Open} & According to the article, what does the \underline{band Four Square hope} to do in the \underline{future}?  \\
 & According to the article we know it is \_ to \underline{prevent} the \underline{forests} from \underline{slowly disappearing}.                            \\ \midrule
\multirow{2}{*}{Amazon-QA}      & For anyone with \underline{small ears}, do these \underline{fit comfortably} or do they \underline{feel} like they are always going to \underline{fall} out, not in correctly, etc. \\ 
  & Does it \underline{remove easily} and does it \underline{leave} any \underline{sticky residue} behind? thanks in advance. \\ 
\bottomrule
\end{tabularx}
\caption{Example of predictions on ARC, RACE-Open and Amazon-QA. Predicted terms are underlined.\label{tbl:es-ex}}
\end{table*}

\subsection{Performance on Open-domain Multiple-choice QA}

We train and evaluate our proposed pipeline method ET-RR on four open-domain multiple-choice QA datasets. All datasets are associated with a sentence-level corpus. Detailed statistics are shown in \Cref{tbl:datasets}. 
\begin{itemize}
\item ARC \citep{Clark2018ThinkYH}: We consider the `Challenge' set in the ARC dataset and use the provided corpus during retrieval.
\item RACE-Open: We adapted the RACE dataset  \citep{lai2017large} to the open-domain setting. Originally, each question in RACE comes with a specific passage. To enable passage retrieval, we concatenate all passages into a corpus with sentence deduplication.\footnote{
As short questions might not contain any words which can relate the question to any specific passage or sentence,
we only keep questions with more than 15 words. 
}
\item MCScript-Open: The MCScript \citep{Ostermann2018SemEval2018T1} dataset is also adapted to the open-domain setting. 
Again we concatenate all passages to build the corpus.\footnote{We keep questions with more than 10 words rather than 15 words to ensure that there is sufficient data. }
\item Amazon-QA: The Amazon-QA dataset \citep{McAuley2016AddressingCA} is an open-domain QA dataset covering over one million questions across multiple product categories. Each question is associated with a free-form answer. We adapt it into a 2-way multiple-choice setting by randomly sampling an answer from other questions as an answer distractor. We split all product reviews at the sentence-level to build the corpus. We consider three categories from the complete dataset in our experiments.
\end{itemize}

In the experiments, ET-RR uses ET-Net to choose essential terms in the question. \Cref{tbl:es-ex} shows example predictions on these target datasets. Then it generates a query for each of the $N$ answer choices by concatenating essential terms and the answer choice. For each query, ET-RR obtains the top $K$ sentences returned by the retriever and considers these sentences as a passage for the reader.
We set $K=10$ for all experiments.

\begin{table*}
\parbox{.55\linewidth}{
\fontsize{9}{11}\selectfont
\centering
\begin{tabular}{lcccc}
\toprule
  Model  &  ARC & RACE-Open & MCScript-Open  \\ 
    & Test & Test & Test  \\
\midrule
IR solver   &  20.26  &  30.70 & 60.46  \\ 
Random   & 25.02 &  25.01 & 50.02   \\
BiDAF     & 26.54 & 26.89 & 50.81  \\
BiLSTM Max-out   & 33.87  & /  & /   \\
ET-RR (Concat)  &   35.33   & 36.87  &  66.46     \\
ET-RR  &   \textbf{36.61}   &  \textbf{38.61} & \textbf{67.71}    \\ 
\bottomrule
\end{tabular}
\caption{Accuracy on multiple-choice selection on ARC, RACE-Open and MCScript-Open.\label{tbl:acc}}
}
\hfill
\parbox{.42\linewidth}{
\fontsize{9}{11}\selectfont
\centering
\vspace{-0.5\baselineskip}
\begin{tabular}{llc}
\toprule
Training Corpus & model  & ARC  \\
\midrule
\multirow{2}{*}{ARC} & Reading Strategies  &  35.0   \\
 &  ET-RR   & 36.6 \\
\midrule
ARC+RACE & Reading Strategies  &  40.7   \\
\bottomrule
\end{tabular}
\caption{
Comparisons of ET-RR and Reading Strategies on ARC.
\label{tbl:arc}}
}
\end{table*}

\begin{table}
\fontsize{9}{11}\selectfont
\vspace{-0.5\baselineskip}
\centering
\vspace{-0.5\baselineskip}
\begin{tabular}{lccc}
\toprule
Model & Amazon & Amazon & Aamzon \\
   &  -Patio &  -Auto  &  -Cell \\
\midrule
IR solver  &  72.80  &  73.60 & 70.50  \\ 
Moqa  &  84.80    &  86.30  &  88.60  \\
ET-RR (Concat)   &  96.19   & 95.21  &  93.26 \\
ET-RR  &   \textbf{96.61}   &  \textbf{95.96} &   \textbf{93.81}   \\ 
\bottomrule
\end{tabular}
\caption{Accuracy on multiple-choice selection on three product categoris of Amazon-QA.\label{tbl:auc}}
\end{table}

We compare ET-RR with existing retrieve-and-read methods on both datasets. As shown in \Cref{tbl:acc}, 
on the ARC dataset, ET-RR outperforms all previous models without using pre-trained models and achieves a relative 8.1\% improvement over the second best BiLSTM Max-out method \citep{mihaylov2018can}. 
Recently, finetuning on pre-trained models has shown great improvement over a wide range of NLP tasks. \citet{Sun2018ImprovingMR} proposed a `Reading Strategies' method to finetune the pre-trained model OpenAI GPT, a language model trained on the BookCorpus dataset \citep{Radford2018ImprovingLU}. They trained Reading Strategies on the RACE dataset to obtain more auxiliary knowledge and then finetune that model on the ARC corpus. 
\Cref{tbl:arc} demonstrates the performance comparison of ET-RR and Reading Strategies on ARC.
As shown, though Reading Strategies trained on both ARC and RACE dataset outperforms ET-RR, ET-RR outperforms Reading Strategies using only the ARC dataset at training time.

\begin{table}
\fontsize{9}{11}\selectfont
\centering
\vspace{-0.5\baselineskip}
\begin{tabular}{lcc}
\toprule
  Pre-trained  &    model      & RACE \\
\midrule
\cmark     & Reading Strategies       & 63.8     \\
\cmark     & OpenAI GPT      &  59.0      \\ \hline
       & ET-RR (reader)           & \textbf{52.3}     \\
       & Bi-attn (MRU)    & 50.4     \\
        & Hier. Co-Matching & 50.4   \\
\bottomrule
\end{tabular}
\caption{Experimental results for reader on RACE.\label{tbl:reader}}
\end{table}

On the RACE-Open and MCScript-Open datasets, ET-RR achieves a relative improvement of 24.6\% and 10.5\% on the test set compared with the second best method IR solver respectively. We also evaluate on multiple categories of the Amazon-QA dataset. As shown in \Cref{tbl:auc}, ET-RR increases the accuracy by 10.33\% on average compared to the state-of-the-art model Moqa \citep{McAuley2016AddressingCA}.
We also compare ET-RR with ET-RR (Concat), which is a variant of our proposed model that concatenates the question and choice as a query for each choice. Among all datasets, ET-RR outperforms ET-RR (concat) consistently which shows the effectiveness
of our essential-term-aware retriever.

\subsection{Ablation study}
We investigate how each component contributes to model performance.

\textbf{Performance of reader.} Our reader alone can be applied on MRC tasks using the
given passages. Here, we evaluate our reader on the original RACE dataset to compare with other MRC models as shown in \Cref{tbl:reader}. As shown,
the recently proposed Reading Strategies and OpenAI GPT models, that finetune generative pre-trained models achieve the highest scores.
Among non-pre-trained models, our reader outperforms other baselines: Bi-attn (MRU) \citep{Tay2018MultirangeRF} and Hierarchical Co-Matching \citep{Wang2018ACM} by a relative improvement of 3.8\%.

\begin{table}
\fontsize{9}{11}\selectfont
\centering
\vspace{-0.5\baselineskip}
\begin{tabular}{lc}
\toprule
Model    & Test \\
\midrule
ET-RR &  \textbf{36.61}  \\ 
-- inter-choice  &  36.36 \\
-- passage-choice &  35.41  \\ 
-- question-choice  &  34.47\\
-- passage-question  & 34.05   \\
\bottomrule
\end{tabular}
\caption{Ablation test on attention components of ET-RR on ARC. `--' denotes the ablated feature.\label{tbl:attention}}
\end{table}

\textbf{Attention components.} \Cref{tbl:attention} demonstrates how the attention components contribute to the performance of ET-RR. As shown, ET-RR with all attention components performs the best on the ARC test set. 
The performance of ET-RR without passage-question attention drops the most significantly out of all the components. It is worth noting that the choice interaction layer gives a further 0.24\% boost on test accuracy.

\textbf{Essential term selection.} To understand the contribution of our essential-term selector, we compare ET-RR with two variants:
(1) ET-RR (Concat) and (2) ET-RR (TF-IDF). For ET-RR (TF-IDF), we calculate the TF-IDF scores and take words with the top 30\% of TF-IDF scores in the question to concatenate with each answer choice as a query.\footnote{According to the annotated dataset, around 30\% of the terms in each question are labelled as essential.}

\begin{table}
\fontsize{9}{11}\selectfont
\centering
\vspace{-0.5\baselineskip}
\begin{tabularx}{\linewidth}{lXXXXXX}
\toprule
\multirow{2}{*}{Model} & \multicolumn{2}{c}{ET-RR}  & \multicolumn{2}{c}{ET-RR}   & \multicolumn{2}{c}{\multirow{2}{*}{ET-RR}}  \\
 & \multicolumn{2}{c}{(Concat)}  & \multicolumn{2}{c}{(TF-IDF)}   &   \\
\midrule
Top $K$ & Dev   & Test   & Dev  & Test  & Dev  & Test   \\
\midrule
5     & 39.26          & 33.36 & 39.93         & 34.73 & 39.93 & 35.59 \\
10    & 38.93          & 35.33 & 39.43         & 35.24 & \textbf{43.96}   &   \textbf{36.61} \\
20    & 41.28          & 34.56 & 38.59         & 33.88 & 42.28 & 35.67 \\
\bottomrule
\end{tabularx}
\caption{Comparison of query formulation methods and amounts of retrieved evidence (i.e.,~top $K$) on the ARC dataset, in terms of percentage accuracy.
\label{tbl:ablation}}
\end{table}

\Cref{tbl:ablation} shows an ablation study comparing different query formulation methods and amounts of retrieved evidence $K$. As shown, with the essential term selector ET-Net, the model consistently outperforms other baselines, given different numbers of retrievals $K$. 
Performance for all models is best when $K=10$. 
Furthermore, only using TF-IDF to select essential terms in a question is not effective.
When $K=10$, the ET-RR (TF-IDF) method performs even worse than ET-RR (Concat). This illustrates the challenges in understanding what is essential in a question. 

Though ET-RR consistently outperforms ET-RR (TF-IDF), the improvement is relatively modest on the Test set (around 
1.4\%). A similar outcome has been reported in \citet{Jansen2017FramingQA, Khashabi2017LearningWI} where essential term extraction methods have shown around 2\%-4\% gain compared with TF-IDF models and struggle to obtain further improvement on SQA tasks. This consensus might show the discrepancy of essential terms between human and machine (i.e.,~the essential terms obtained using a human annotated dataset might not be helpful in a machine inference model). Another reason might be the current retrieval method does not effectively use these essential terms and the performance highly depends on the dataset. Note that the ET-RR outperforms ET-RR (TF-IDF) by around 4\% on the Dev set. Therefore, how to develop well-formed single or even multi-hop queries using these terms are worth studying in the future.

\subsection{Error Analysis}

\Cref{tbl:arc-example} shows two major types of error, where the correct answer choice is in \textbf{bold} and the predicted answer choice is in \textit{italics}.

\noindent\textbf{Retrieved supporting evidence but failed to reason over it.} For the first question, there exists evidence that can justify the answer candidate (C). However, the model chooses (D) which has more words overlapping with its evidence. This shows that the model still lacks the reasoning capability to solve complex questions.

\noindent\textbf{Failed to retrieve supporting evidence.} For the second question, the retrieved evidence of both the correct answer (D) and the prediction (B) is not helpful to solve the question. Queries such as `what determines the year of a planet' are needed to acquire the knowledge for solving this question. This poses further challenges to design a retriever that can rewrite such queries.

\begin{table}
\parbox{\linewidth}{
\fontsize{9}{11}\selectfont
\centering
\vspace{-0.5\baselineskip}
\begin{tabularx}{\linewidth}{X}
\toprule
The elements carbon, hydrogen, and oxygen are parts of many different compounds. Which explains why these three elements can make so many different compounds? \\ \hline
(A) They can be solids, liquids, or gases. \\
(B) They come in different sizes and shapes. \\
\textbf{(C) They combine in different numbers and ratios.} \\
* There are many different types of compounds because atoms of elements combine in many different ways (and in different whole number ratios) to form different compounds. \\
\textit{(D) They can be a proton, a neutron, or an electron.} \\
* Atoms of different elements have a different number of protons, neutrons, and electrons. \\
\midrule
\midrule
Which planet in the solar system has the longest year? \\ \hline
(A) The planet closest to the Sun. \\
\textit{(B) The planet with the longest day.} \\
* The planet with the longest day is Venus; a day on Venus takes 243 Earth days. \\
(C) The planet with the most moons. \\
\textbf{(D) The planet farthest from the Sun.} \\
* The last planet discovered in our solar system is farthest away from the sun. \\
\bottomrule
\end{tabularx}
}
\caption{Examples where ET-RR fails on ARC. The retrieved evidence for each answer candidate is marked by *.}
\label{tbl:arc-example}
\end{table}

\section{Conclusion}
We present a new retriever-reader model (ET-RR) for open-domain QA.
Our pipeline has the following contributions:
(1) we built an essential term selector (ET-Net) which helps the model understand which words are essential in a question leading to more effective search queries when retrieving related evidence;
(2) we developed an attention-enhanced reader with attention and fusion among passages, questions, and candidate answers. Experimental results show that ET-RR outperforms existing QA models on open-domain multiple-choice datasets as ARC Challenge, RACE-Open, MCScript-Open and Amazon-QA. 
We also perform in-depth error analysis to show the limitations of current work. For future work, we plan to explore the directions of (1) constructing multi-hop query and (2) developing end-to-end retriever-reader model via reinforcement learning.

\noindent\textbf{Acknowledgements.} We thank Jade Huang for proofreading the paper, Liang Wang and Daniel Khashabi for sharing code and the annotated dataset with us. This work is partly supported by NSF \#1750063. We thank all the reviewers for their constructive suggestions.

\bibliography{naaclhlt2019}
\bibliographystyle{acl_natbib}



\end{document}